\documentclass{isprs}
\usepackage{natbib}
\usepackage{subfig}
\usepackage{float}
\usepackage{setspace}
\usepackage{geometry}
\usepackage{epstopdf}
\usepackage{siunitx}
\usepackage[hidelinks]{hyperref}
\usepackage[labelsep=period]{caption}
\usepackage[british]{babel} 
\usepackage[hang]{footmisc}

\begin{document}

\title{SEMANTIC SEGMENTATION OF URBAN TEXTURED MESHES THROUGH POINT SAMPLING}
\date{}

\author{Grégoire Grzeczkowicz\textsuperscript{1, 2,}\thanks{Corresponding author} , Bruno Vallet \textsuperscript{1}}
\address{\textsuperscript{1} LASTIG, Univ Gustave Eiffel, IGN, ENSG \\ \textsuperscript{2} Direction Générale de l'Armement \\ \href{mailto:Gregoire.Grzeczkowicz@ign.fr}{\texttt{Gregoire.Grzeczkowicz@ign.fr}}}

\icwg{}   

\abstract{Textured meshes are becoming an increasingly popular representation combining the 3D geometry and radiometry of real scenes. However, semantic segmentation algorithms for urban mesh have been little investigated and do not exploit all radiometric information. To address this problem, we adopt an approach consisting in sampling a point cloud from the textured mesh, then using a point cloud semantic segmentation algorithm on this cloud, and finally using the obtained semantic to segment the initial mesh. In this paper, we study the influence of different parameters such as the sampling method, the density of the extracted cloud, the features selected (color, normal, elevation) as well as the number of points used at each training period. Our result outperforms the state-of-the-art on the SUM dataset, earning about 4 points in OA and 18 points in mIoU.}


\maketitle

\section{INTRODUCTION}\label{INTRODUCTION}

\sloppy

Textured 3D meshes of urban areas are becoming more and more common and of ever better quality. A mesh defines a continuous 3D surface (2-manifold) through a set of triangles sharing edges and vertices, which representation is more appropriate than its usual alternatives:
\begin{itemize}
    \item Digital Surface Models (DSM) represent the geometry of the scene by an elevation ($z$) sampled on a regular horizontal grid. This representation is simple and efficient but discards information on vertical surfaces (walls, facades,~...) and overhangs (bridges, tunnels,~...).
    \item Point clouds simply consist of discrete samples of the geometry without recovering the continuous nature of the real surface.
\end{itemize}
Meshes are light and adaptive data structures where large areas can be represented by few large faces while details can be captured by a larger number of small faces. Meshes can be textured by creating a bijection between subsets of faces and their projection in 2D images called textures \citep{waechter2014}, allowing them to hold a radiometric information in the same way that DSMs (a color per pixel) and point clouds (a color per point). Textured meshes are now becoming the standard output of the photogrammetric reconstruction pipeline, as the preferred way to gather radiometric and geometric data of an urban scene \citep{laupheimer2021}. Last but not least, textured meshes are the main representation used by the video games and animation industries and numerous techniques have been developed to process and visualize them. For all these reasons the semantic segmentation of textured meshes have received a growing interest. While state of the art algorithms currently rely on random forests and Markov Random Fields, the most recent algorithms leverage graph convolutions and deep neural networks. Semantic segmentation consists in associating to each piece of data the semantic class to which it belongs (for example, in the urban context, \textit{buildings}, \textit{ground} or \textit{vegetation}).
While these pieces of data are natural for DSMs (pixels) and point clouds (points), different choices can be made for textured meshes: vertices, triangles or texels (pixels of the texture images), which will be discussed in the paper.
Semantic segmentation of urban meshes raises open questions in terms of adapted network architectures for learning and poses some important challenges in terms of scaling. Moreover, to the best of our knowledge, radiometric information is only weakly exploited at the moment, most algorithms simply extracting features from the radiometric information per face, rather than using the whole texture. In this paper, we propose to use the textured mesh to sample a point cloud of very good quality with colors and normal in order to use a high-performance semantic segmentation algorithm on point clouds. We investigate different techniques for sampling points on a textured mesh and compare the results obtained using the KPConv algorithm \citep{thomas2019} on the resulting point clouds.

\section{RELATED WORK}\label{sec:RELATED WORK}

\sloppy

\subsection{Mesh Semantic Segmentation}\label{sec:Mesh Semantic Segmentation}

The pioneer works on mesh semantic segmentation were performed using handcrafted features (geometrical and radiometric) with a Random Forest classifier and Markov Random Fields \citep{rouhani2017}. The selected features are elevation, planarity, verticality, average color, standard deviation, and color distribution in the HSV color space. \citet{tutzauer2019} use a 1x1 CNN to learn one features vector per face from multi-scale features aggregation from neighbors faces. These learned features are combined with handcrafted features to feed a Random Forest classifier. \citet{laupheimer2020} use the same approach to compute per face features vectors, but then use PointNet++ \citep{qi2017b} to segment the mesh view as a point cloud where each face is a point.

As a mesh represents a 2D surface, it is a 2-manifold which locally resembles (2D) Euclidean space. This characteristic can be exploited to use standard CNNs, by locally parameterize the surface to 2D \citep{masci2015,maron2017}. \citet{tatarchenko2018} introduce tangent convolution, where a small neighborhood around each point is used to reconstruct the local function upon which convolution is applied.

As the structure of a mesh carries two graphs (the primal, formed by vertices connected by edges and the dual, formed by faces connected by dual edges), classical Graph Convolutional Neural Networks can be adapted to work on a mesh. \citet{verma2018} implement the pooling operation through a generic graph clustering algorithm \citep{dhillon2007}. \citet{ranjan2018} use classic mesh-simplification techniques for surface approximation \citep{garland1997} to define their pooling operator. \citet{milano2020} designed a primal-dual framework GCN named PD-MeshNet. PD-MeshNet takes the edges and faces of the 3D meshes as the features of the graph nodes, and then adds an attention mechanism to achieve classification and segmentation. MeshCNN \citep{hanocka2019} introduce a mesh-specific convolution and pooling layers that are applied over the edges of a mesh. \citet{knott2021} add radiometric features to face inspired by \citet{rouhani2017}.

\subsection{Point Cloud Semantic Segmentation}\label{sec:Point Cloud Semantic Segmentation}

Several approaches are used to encode point clouds for semantic segmentation. The simplest one is to use voxels (cubic volume elements of a regular 3D grid). The interest is that voxels generalize the notion of pixel to 3D, thus allowing for a trivial extension of Convolutional Neural Networks (CNNs).
The first to use this approach are \citet{wu2015} with a CNN that performs shape classification. Even if this approach is theoretically simple, it has a very large memory footprint because a 3D point cloud is very sparse when discretized on voxels. Different approaches have been used to exploit this sparsity to use reduced space representation (FPNN \citep{li2016}, OctNet \citep{riegler2017}, O-CNN \citep{wang2017}, SparseConvNet \citep{graham2018}).

Recent progress has been made to use a point cloud directly as input to a neural network. \citet{qi2017a} designed PointNet, a solution to the point order invariance problem that uses a 1x1 convolution followed by global max pooling. PointNet++ \citep{qi2017b} introduces a partition of points to better consider local structures. This approach has inspired many sophisticated neural modules to learn per-point local features. These modules can be generally classified as neighboring feature pooling (So-net \citep{li2018}, PointWeb \citep{zhao2019}, ShellNet \citep{zhang2019a}), graph message passing (LocalSpecGCN \citep{wang2018}, GAC \citep{wang2019b}, EdgeConv \citep{wang2019a}, ClusterNet \citep{chen2019}), kernel-based convolution (SPLATNet \citep{su2018}, PointConv \citep{wu2019}, A-CNN \citep{komarichev2019}, KPConv \citep{thomas2019}, LatticeNet \citep{rosu2020}) and attention-based aggregation (ShapeContextNet \citep{xie2018}, PCAN \citep{zhang2019b}, GSS \citep{yang2019}).

Nevertheless, most of these approaches are often unsuitable for processing the large-scale point clouds arising from high resolution LiDAR scans or dense matching on urban areas. Super Point Graph (SPG) \citep{landrieu2018} introduces the notion of superpoints, an equivalent of superpixels for point clouds to partition the point cloud into subsets of points that share the same semantic class, then semantize the resulting SPG which size is much more reasonable than the initial point cloud with graph convolutions. RandLA-Net \citep{hu2020} use both random point sampling (for memory and computation efficiency) and a novel local feature aggregation module that progressively increases the receptive field for each point (for preserving geometric details). These approaches allow the processing of much larger point clouds.

\section{METHOD}\label{sec:METHOD}

\sloppy

\subsection{Mesh Sampling with points}\label{sec:Point Sampling on Mesh}

While for image and point cloud semantic segmentation, the ``piece of data'' to semantize is natural (pixels and points), the question is more open for textured meshes:  do we look for a semantic labelling of vertices, triangles or texels ? In fact we propose another alternative: semantizing discrete points sampling the mesh. This has multiple benefits:
\begin{itemize}
    \item We can leverage the abundant literature on point cloud semantic segmentation on the resulting samples
    \item By sampling more or less points, we can control the performance vs quality compromise.
    \item By sampling points regularly we can overcome typical limitations of point cloud semantic segmentation that are sensitive to sampling anisotropy and strongly varying point densities.
    \item The mesh gives an unambiguous normal at each point.
    \item The texture gives a color at each point.
    \item The labels obtained per point samples can easily be interpolated to the vertices, faces or texels.
\end{itemize}
In this paper, we study two types of sampling: texel sampling based on the texture and a random but homogeneous sampling called Poisson disk sampling.

\begin{figure}[b]
\begin{center}
	\includegraphics[width=1.0\columnwidth]{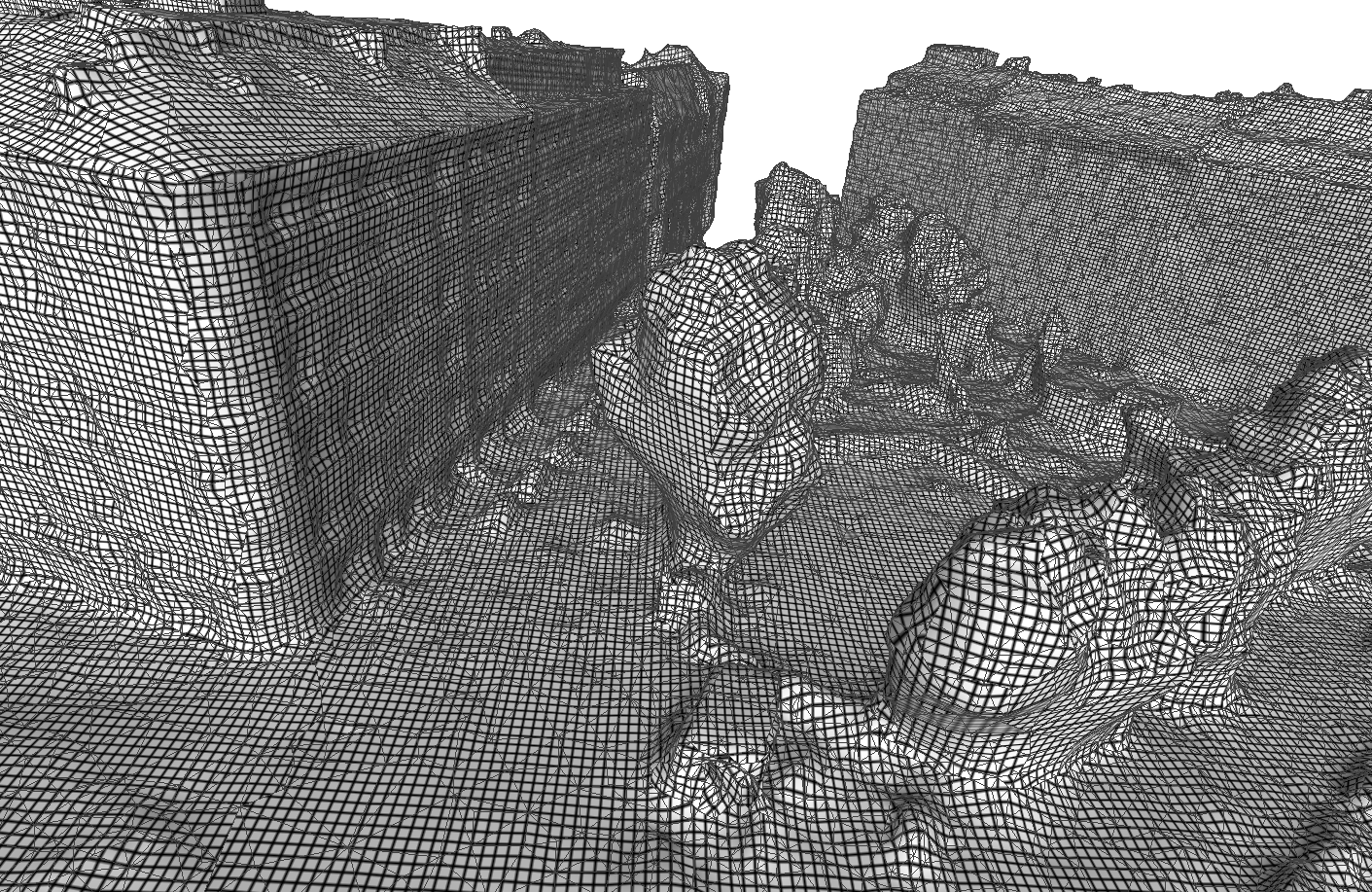}
	\caption{Texture grid on the mesh.}
\label{fig:grid_visualisation}
\end{center}
\end{figure}

A first way to obtain a homogeneous sampling is to sample the points on a regular grid defined on the surface of the mesh. For this we can take advantage of the texture of our mesh which defines a natural grid globally homogeneous on the surface of the mesh \citep{waechter2014}. This texture grid has the advantage of being continuous between some faces that share the same texture patch (Figure \ref{fig:grid_visualisation}). Sampling one point at the center of each texel, which we call texel sampling, is moreover perfectly adapted to the quantity of texture information as each pixel is converted into a point (Figure \ref{subfig:pxl-1.0}). By adapting the resolution of the texture used, we can choose the density of the point cloud produced (Figure \ref{subfig:pxl-2.0}). We have introduced a parameter $s$ which represents the ratio of the retained pixel size to the original size.

\begin{figure}[H]
 \begin{center}
    \subfloat[Mesh]{
        \includegraphics[width=0.485\columnwidth]{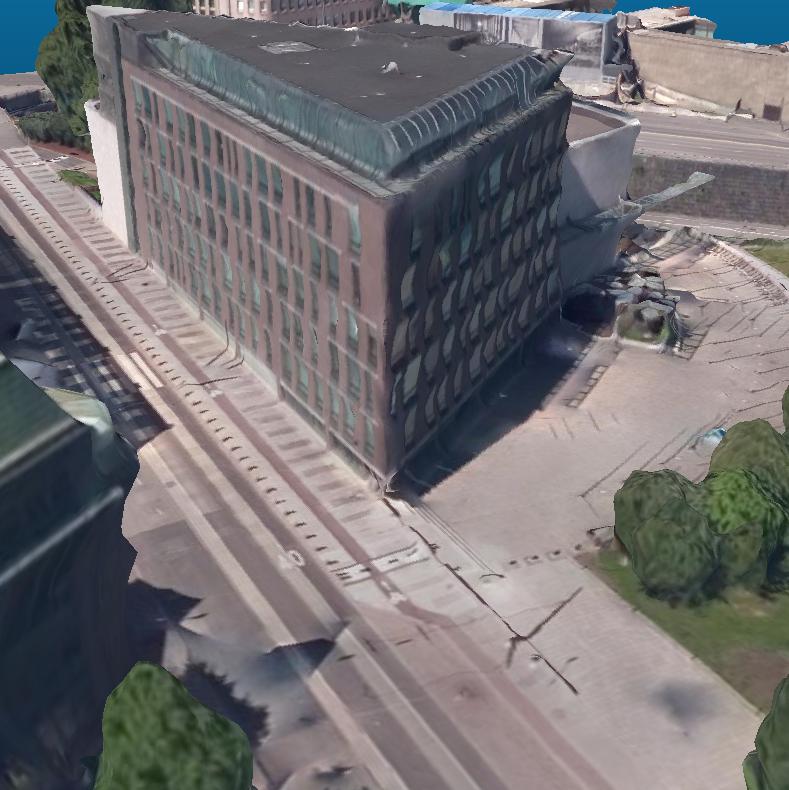}
        \includegraphics[width=0.485\columnwidth]{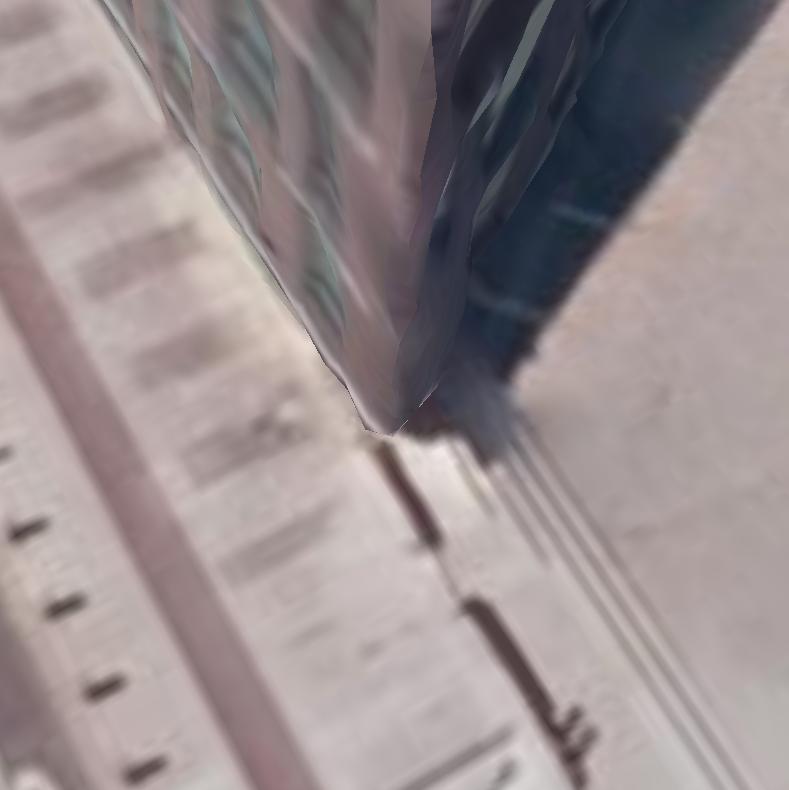}
        \label{subfig:mesh}
    }
    \hspace{0pt}
    \subfloat[Texel sampling ($s = 1.0$)]{
        \includegraphics[width=0.485\columnwidth]{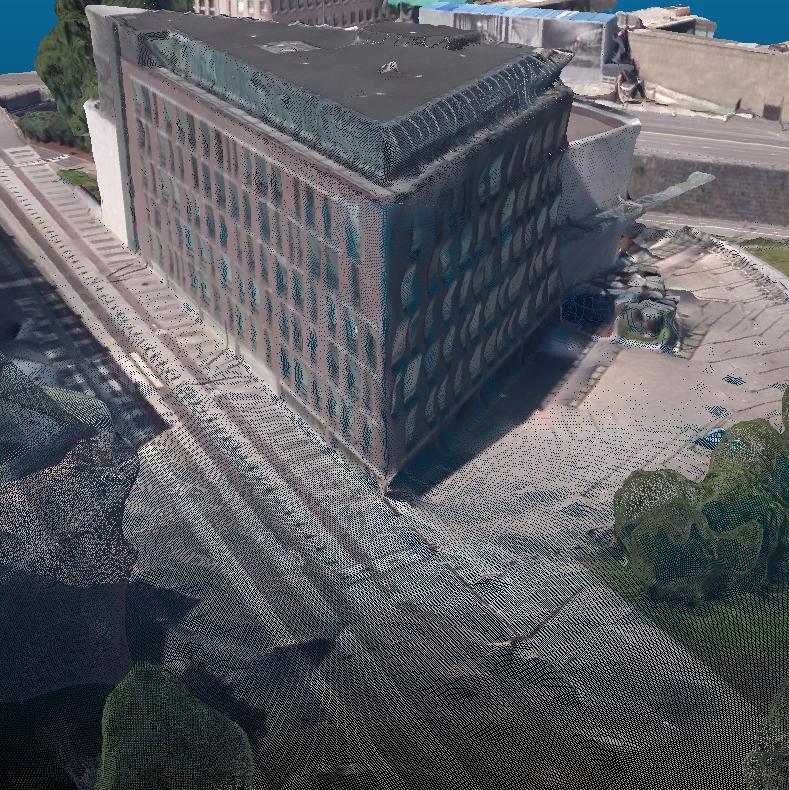}
        \includegraphics[width=0.485\columnwidth]{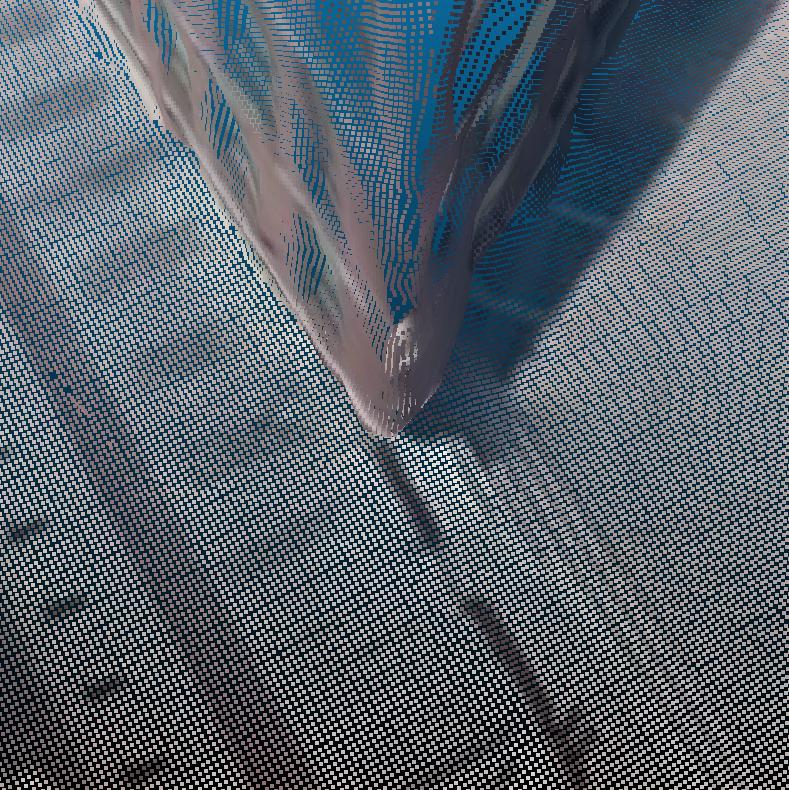}
        \label{subfig:pxl-1.0}
    }
    \hspace{0pt}
    \subfloat[Texel sampling ($s = 2.0$)]{
        \includegraphics[width=0.485\columnwidth]{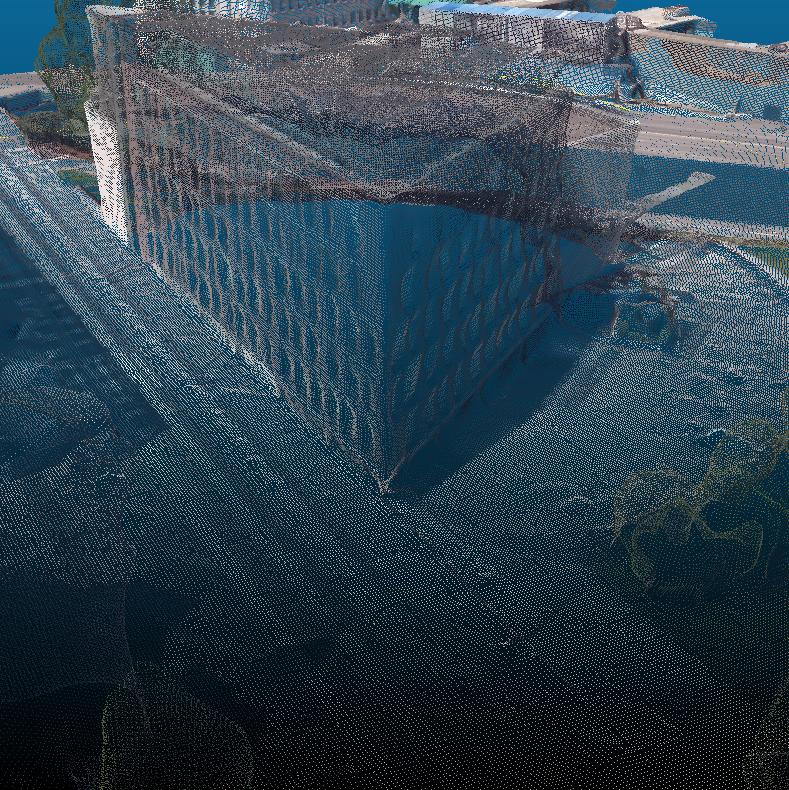}
        \includegraphics[width=0.485\columnwidth]{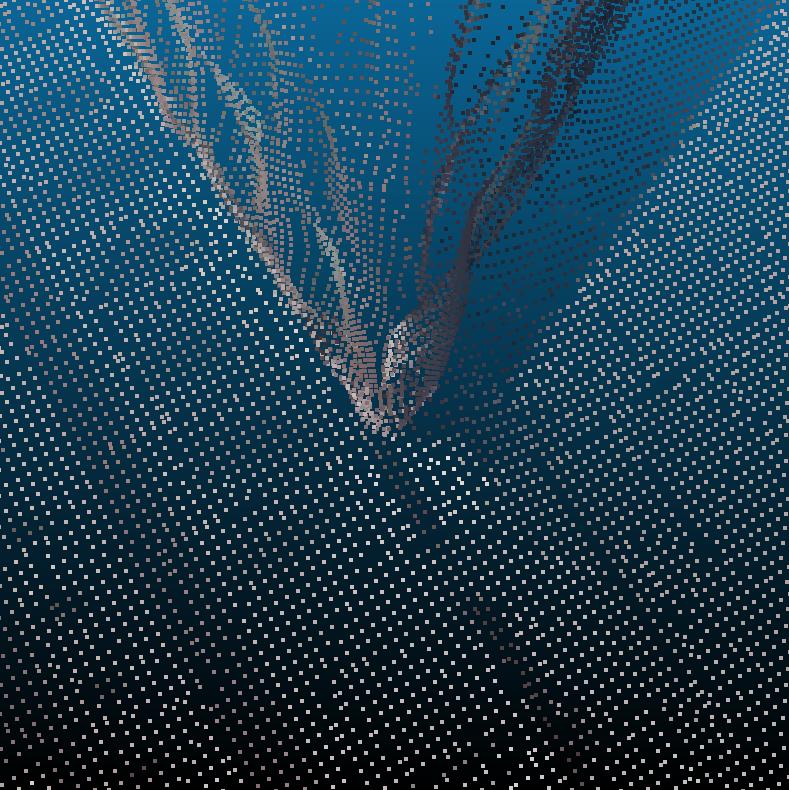}
        \label{subfig:pxl-2.0}
    }
    \hspace{0pt}
    \subfloat[Texel sampling ($s = 1.0$) + grid sub-sampling ($g = 0.2$)]{
        \includegraphics[width=0.485\columnwidth]{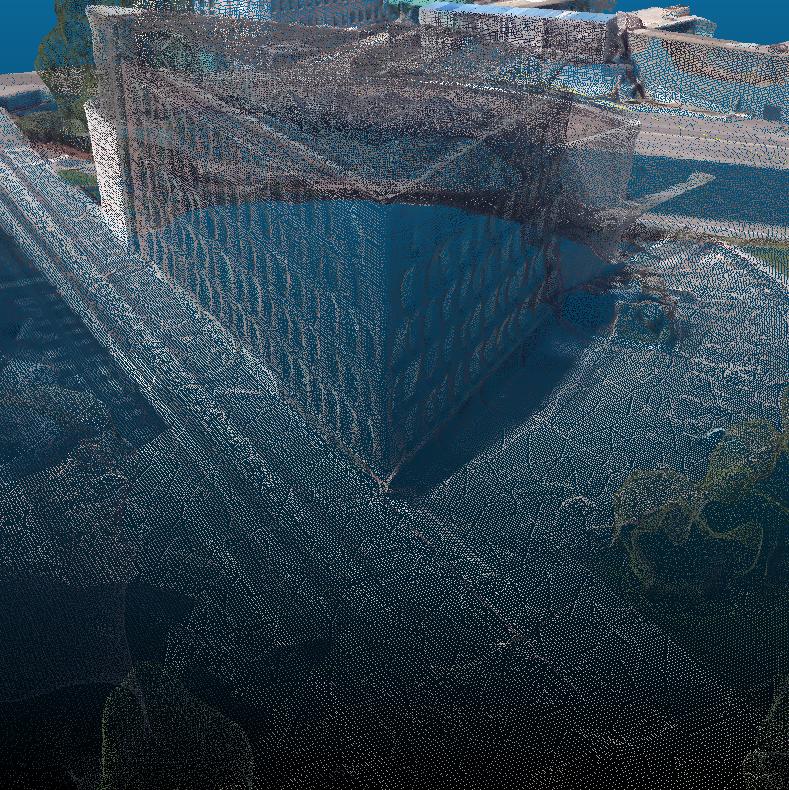}
        \includegraphics[width=0.485\columnwidth]{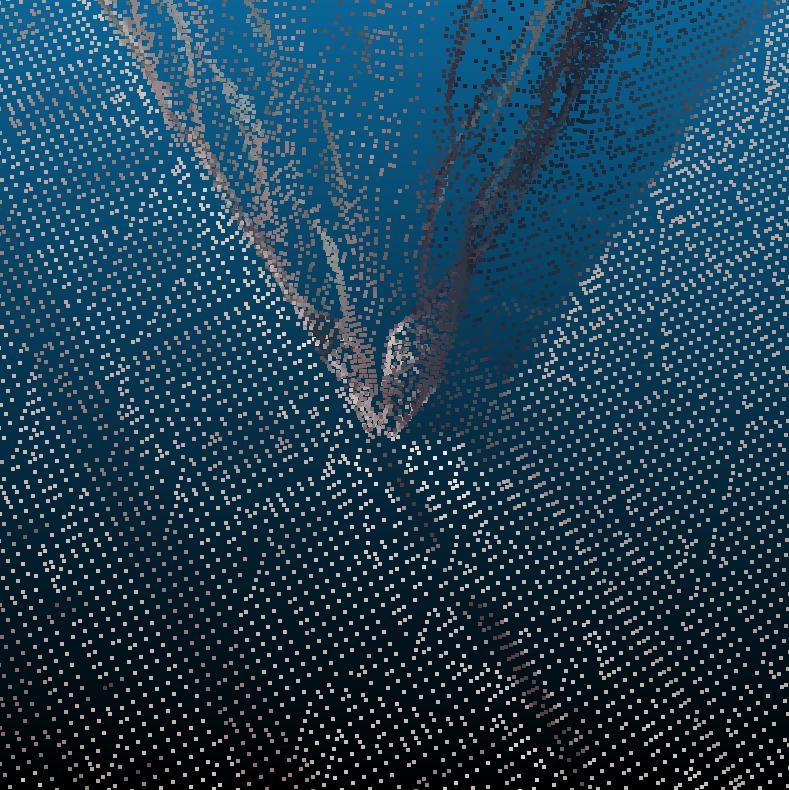}
        \label{subfig:pxl-1.0-gs-0.2}
    }
\caption{Sampling methods comparison.}
\label{fig:sampling comparison}
\end{center}
\end{figure}

A second approach is to perform a random but homogeneous sampling on the surface. A well known solution to this problem is the Poisson disk sampling algorithm. A Poisson disk sampling is a set of points such that all samples are at least at a distance $r$ apart for some user-supplied density parameter $r$ (Figures \ref{subfig:pds-0.1} and \ref{subfig:pds-0.15}). The naïve rejection-based approach for generating Poisson disk samples, dart throwing, is impractically inefficient and requires a stop condition. Over the years, a large number of methods for generating Poisson disk distributions have been proposed \citep{lagae2008}. We choose to use the Constrained Sample-based Poisson disk Sampling developed specifically for meshes and integrated in Meshlab \citep{corsini2012}. The main idea of the algorithm is to sample the mesh with uniformly random points and then to subsample this set of points by choosing the points to keep one by one while removing from the set the points whose distance to the previously chosen points is less than $r$. The sampling stops when all the initial points have been either chosen or removed.

\begin{figure}[H]
\ContinuedFloat
\centering
    \subfloat[Poisson disk sampling ($r = 0.1$)]{
        \includegraphics[width=0.485\columnwidth]{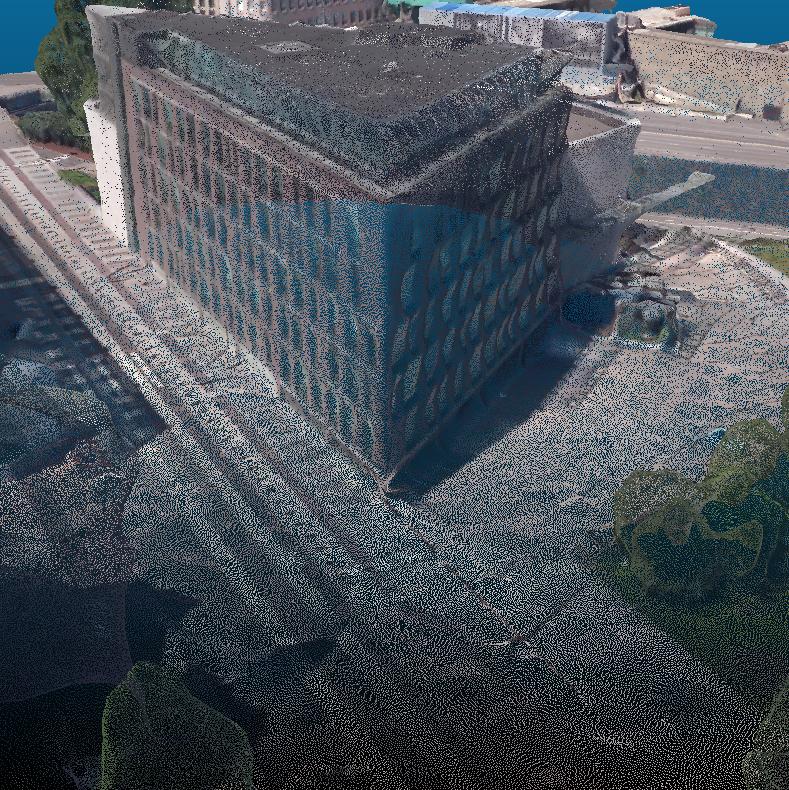}
        \includegraphics[width=0.485\columnwidth]{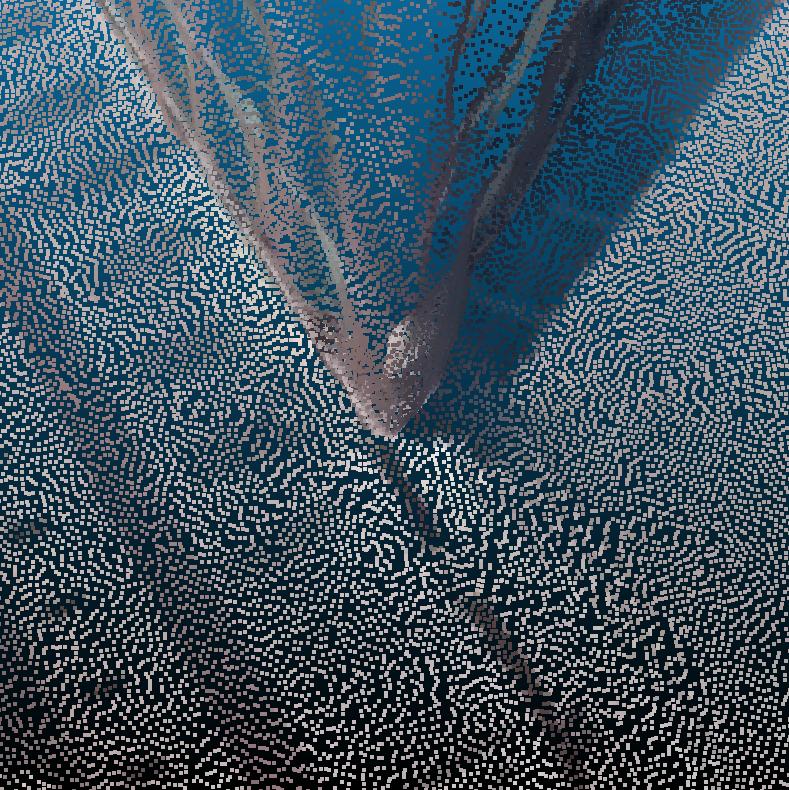}
        \label{subfig:pds-0.1}
    }
    \hspace{0pt}
    \subfloat[Poisson disk sampling ($r = 0.15$)]{
        \includegraphics[width=0.485\columnwidth]{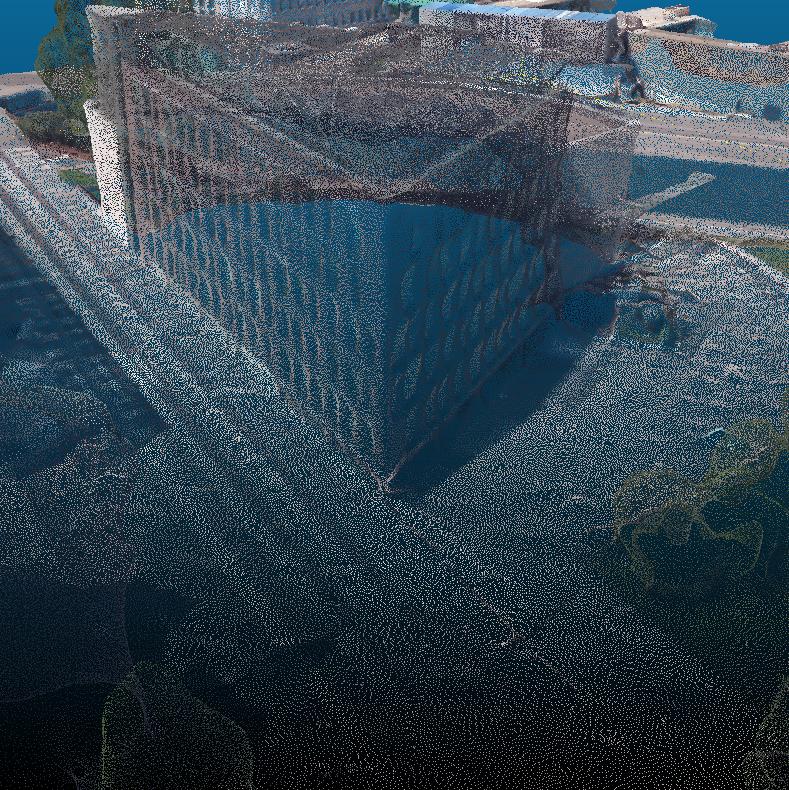}
        \includegraphics[width=0.485\columnwidth]{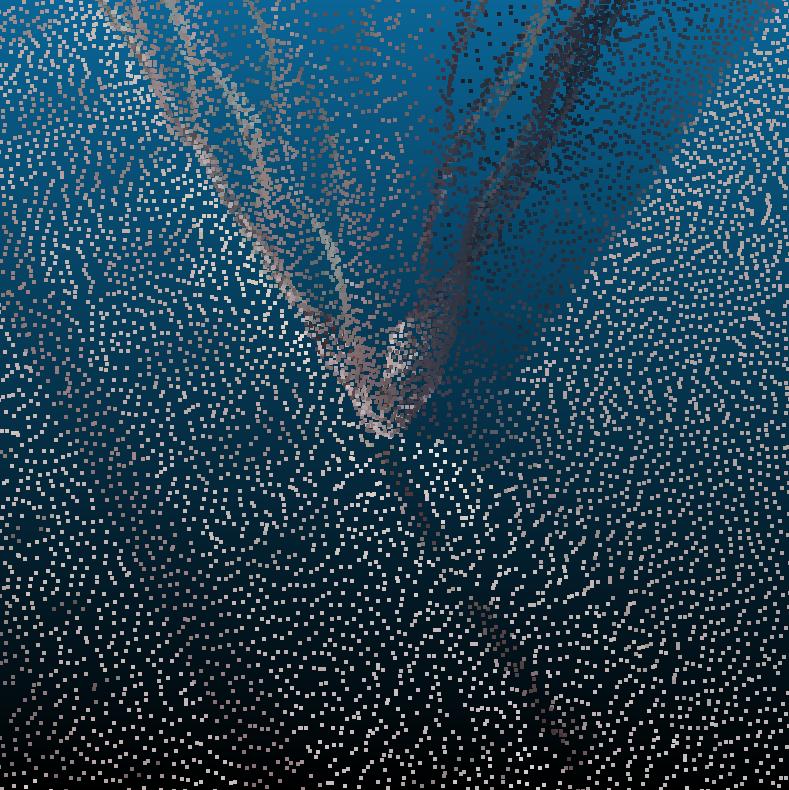}
        \label{subfig:pds-0.15}
    }
    \hspace{0pt}
    \subfloat[Poisson disk sampling ($r = 0.1$) + grid sub-sampling ($g = 0.2$)]{
        \includegraphics[width=0.485\columnwidth]{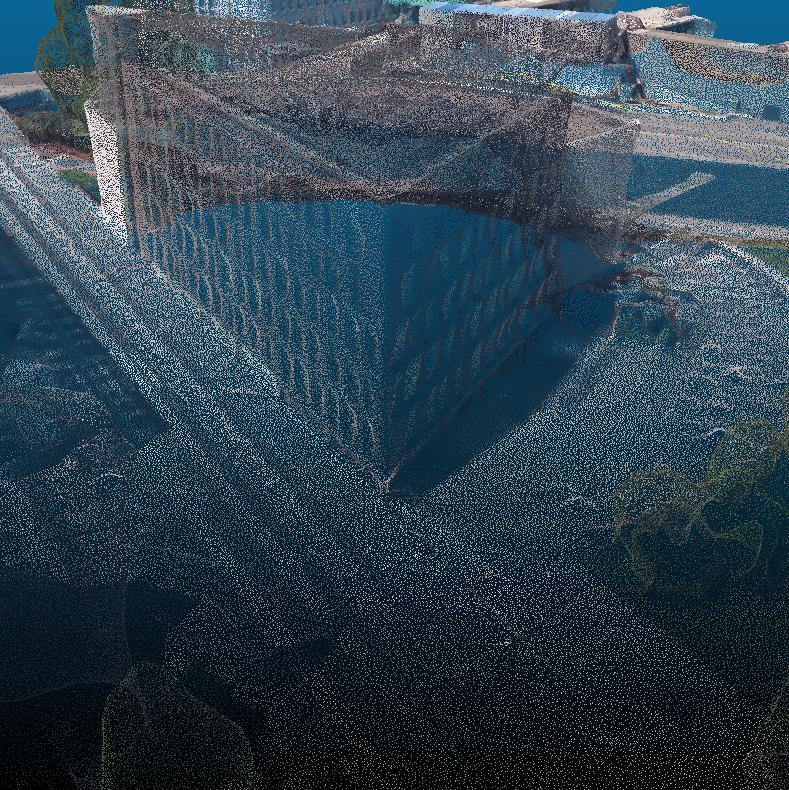}
        \includegraphics[width=0.485\columnwidth]{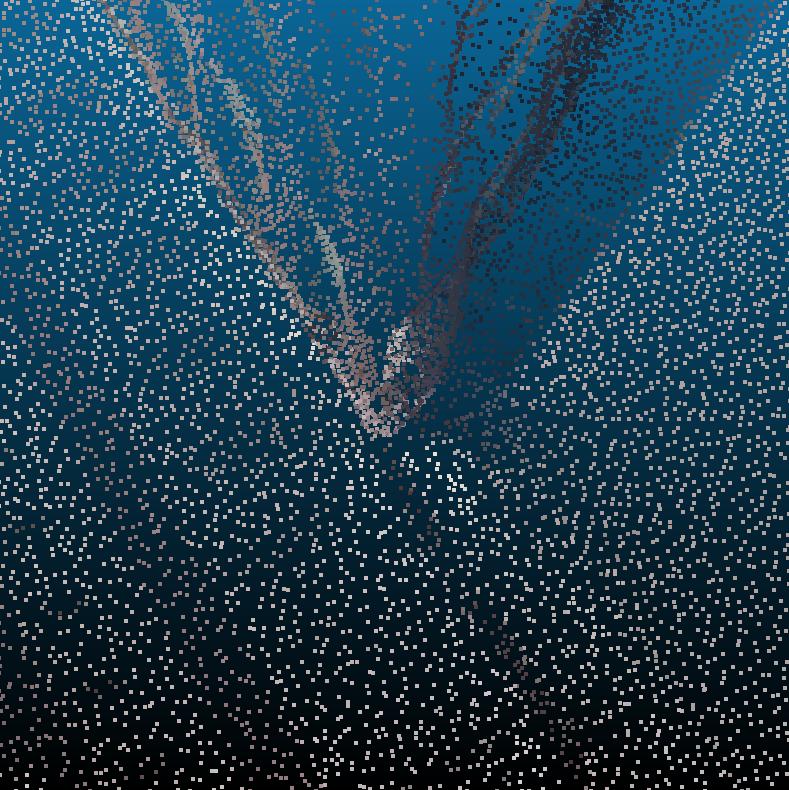}
        \label{subfig:pds-0.1-gs-0.2}
    }
\caption{Sampling methods comparison.}
\label{fig:sampling comparison 2}
\end{figure}

\begin{figure*}[t]
\begin{center}
	\includegraphics[width=0.7\textwidth]{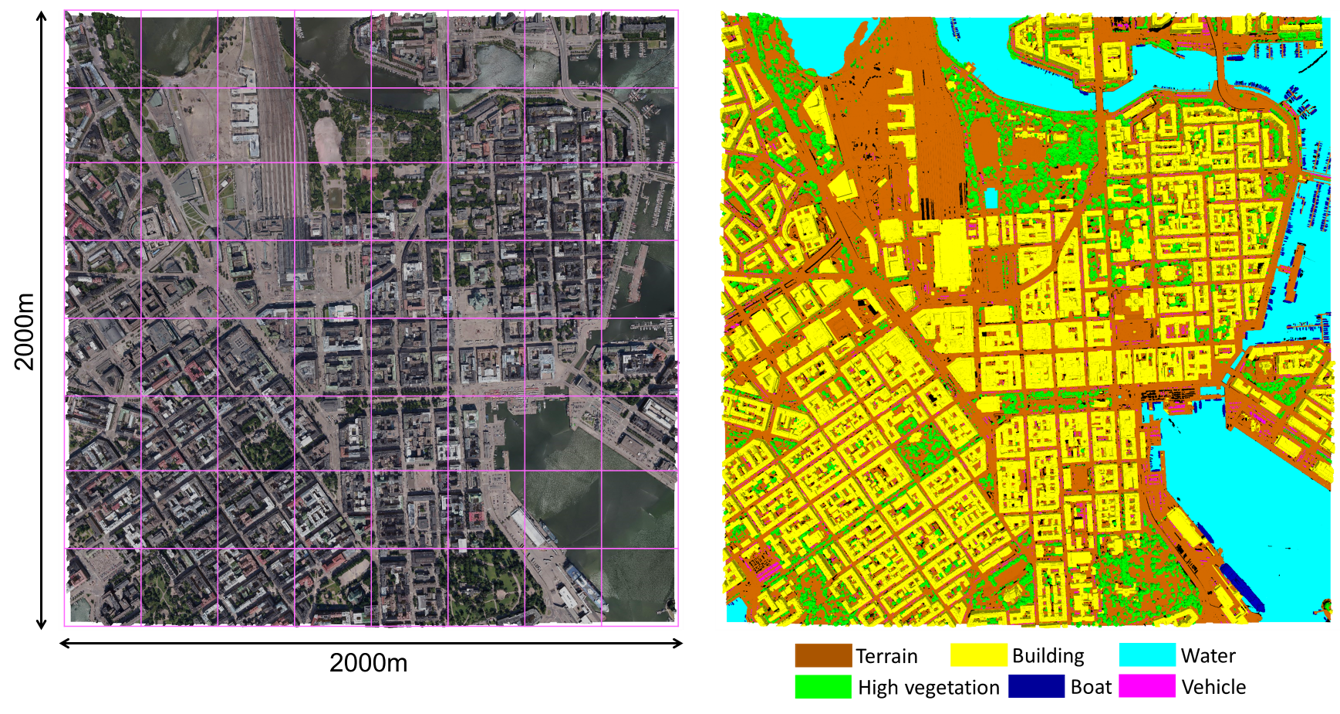}
	\caption{SUM dataset \protect\citep{gao2021}.}
\label{fig:dataset_example}
\end{center}
\end{figure*}

Whether it is pixel sampling or Poisson disk sampling, these two methods allow to adapt the density of the point cloud produced (by adapting the resolution of the texture used or by adapting the distance $r$). Nevertheless, the density of the produced cloud can also be adapted by performing a sub-sampling after the cloud generation. \citet{hu2021} has compared grid sub-sampling and random sub-sampling and obtained better results with grid sub-sampling, thus we tested grid sub-sampling (Figures \ref{subfig:pxl-1.0-gs-0.2} and \ref{subfig:pds-0.1-gs-0.2}). By choosing the grid step $g$, we can adapt the density of the output point cloud. We have chosen to randomly keep one point per cell of the grid, rather than generating a centered point per cell that would average the points in the cell. This allows us to obtain a greater geometric fidelity to the initial mesh. If the initial sampling is not random, an aliasing effect may appear after sub-sampling, as in our case with texel sampling (Figure \ref{subfig:pxl-1.0-gs-0.2}).

\subsection{Feature selection}\label{sec:Feature selection}

The textured mesh sampling allows us to associate multiple input features to each point. We have studied the influence on the segmentation of several features such as the color, normal or the elevation. For the normal, we propose two alternatives: the face normal or an interpolation of the vertices normal. The face normal is directly the normal of the triangle on which the point is sampled, oriented towards the outside of the mesh. The normal at a vertex of a mesh is usually defined as the average of the normals of the adjacent faces weighted by their area. If this sum is null, we defined the normal of the vertex as the sum of the director vectors of the edges adjacent to this vertex, as these cases occur when two faces of the mesh fold in on themselves. In both cases, this vector is normalized to obtain a unit vector. The normal of a point is then defined as the average of the normals at the vertices of the face to which it belongs, weighted by the inverse distance to these vertices. The face normals are equal for all points on the same face, thus discontinuous between faces, while the interpolated normals are continuous even at the edges and vertices.

To calculate the elevation of a point, we can use two methods. The first is simpler, but less accurate. It consists of calculating the difference between the elevation of the point and the minimum elevation of the points contained in a neighborhood large enough to contain ground and small enough that the ground can be considered as globally flat in this neighborhood. The difficulty is to define this neighborhood, if it exists, and that sometimes points may be located below the surface of the ground (for example in the case of a river). The second method consists of deriving a digital terrain model (DTM) from the mesh \citep{beumier2016} and then determining the elevation of each point with respect to this model. The main problem with this method is that deriving a DTM from the mesh is a non-trivial problem. We therefore chose in the case of our study to test only the elevation obtained using the first method.

\section{EVALUATION}\label{sec:EVALUATION}

\sloppy

\subsection{Dataset}\label{sec:Dataset}

While a large number of textured meshes of urban environments are publicly available, very few have been labeled to serve as training, validation and test data for semantic segmentation. Indeed, urban environment meshes are often quite large (they can contain from one to several tens of millions of faces according to the covered surface) and thus require many hours of work to be labelled manually. To evaluate our different semantic segmentation methods, we used the SUM dataset \citep{gao2021} which is a textured mesh of the city of Helsinki covering an area of \SI{4}{\kilo\metre\squared} (Figure \ref{fig:dataset_example}). It is a semi-automatic labeling of the raw Helsinki 3D dataset \citep{helsinki2017}: a textured mesh that was generated in 2017 with ContextCapture \citep{contextcapture2016} from oblique aerial images with a ground sampling distance (GSD) around 7.5 cm which covers \SI{12}{\kilo\metre\squared}.

\begin{figure}[ht]
\begin{center}
	\includegraphics[width=0.6\columnwidth]{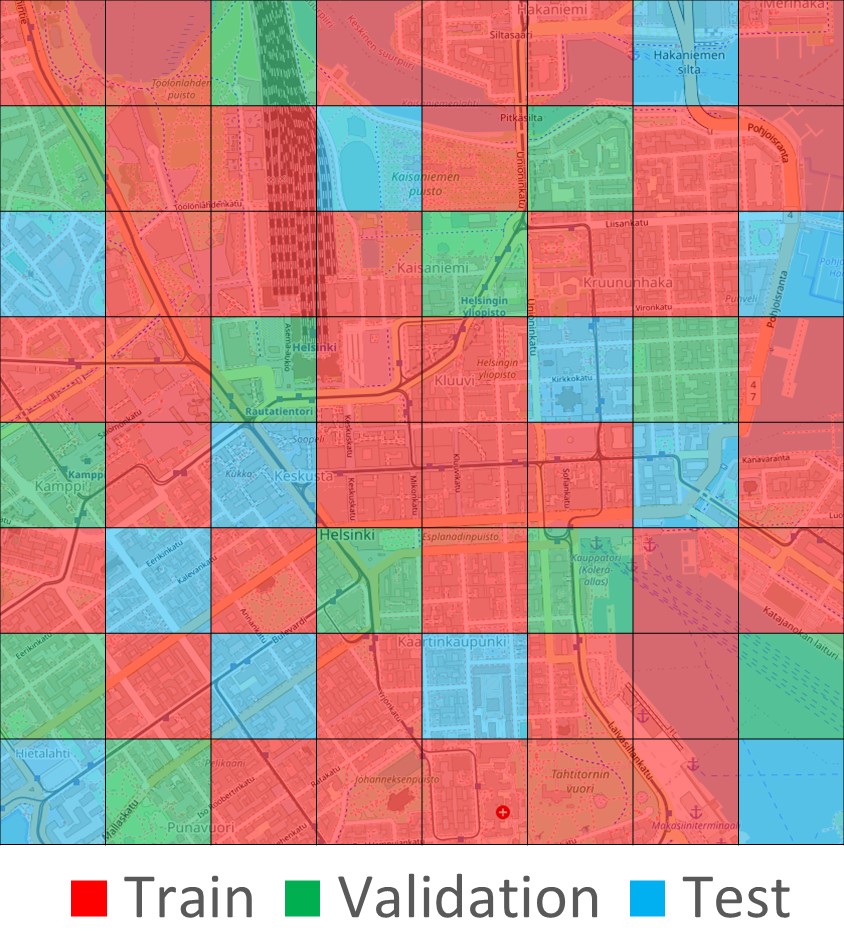}
	\caption{Tile distribution between set.}
\label{fig:tile_distribution}
\end{center}
\end{figure}

The source images have three colour channels (i.e., red, green, and blue) and the triangular faces are divided into 6 categories: terrain, vegetation, building, water, vehicle and boat. Ambiguous regions (which account for about \SI{2.6}{\percent} of the total mesh surface area), such as shadowed regions or distorted surfaces, are labelled as unclassified. The entire mesh is split into 64 tiles, and each of them covers about \SI{250}{\metre\squared}. 40 randomly selected tiles are used as training data, 12 as validation data and 12 as test data (Figure \ref{fig:tile_distribution}).

\begin{figure}[ht]
\begin{center}
	\includegraphics[width=1.0\columnwidth]{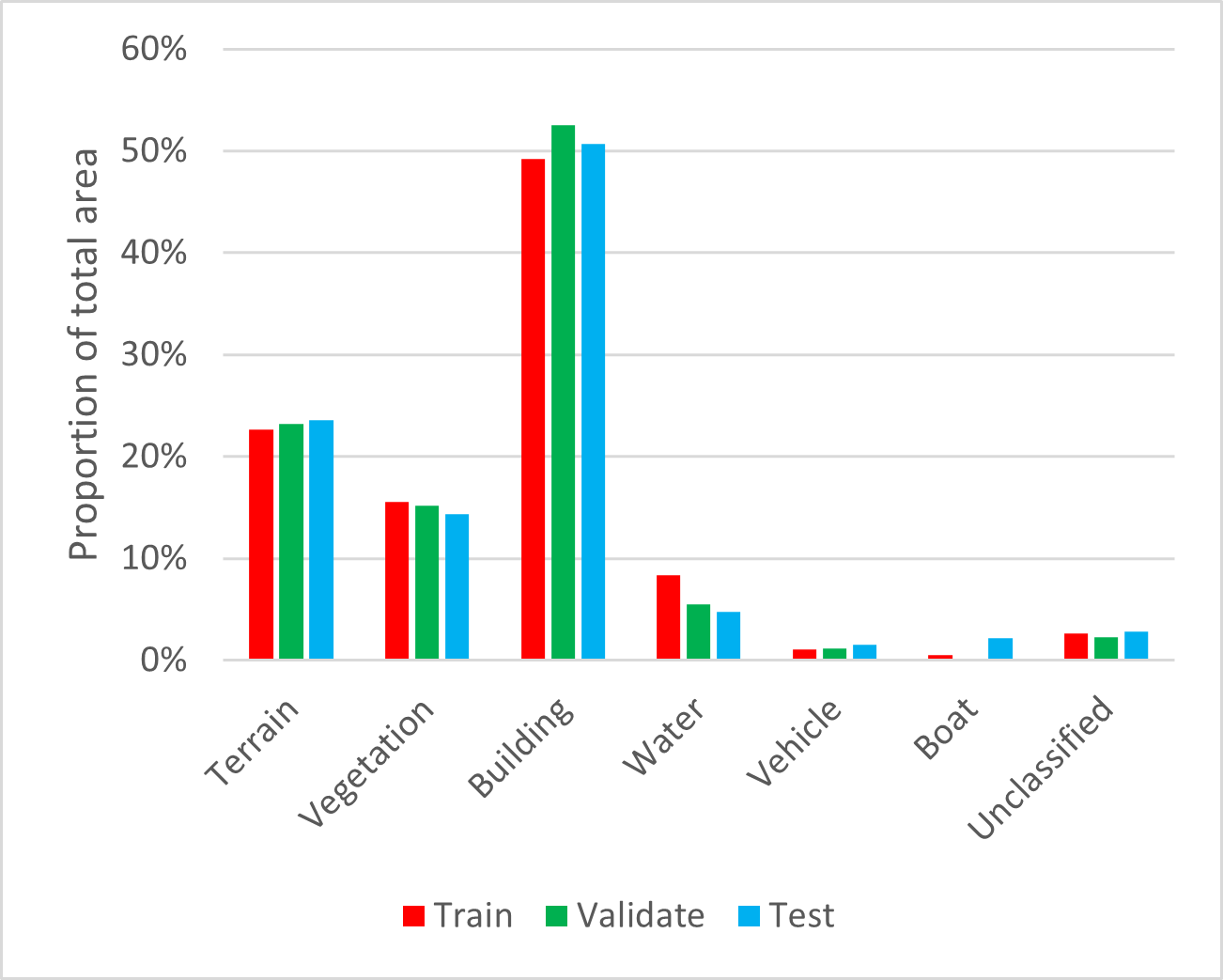}
	\caption{Surface distribution between classes for each set.}
\label{fig:surface_distribution}
\end{center}
\end{figure}

For each of the seven semantic categories, we computed the total area in the training, validate and test dataset to show the class distribution. As shown in Figure \ref{fig:surface_distribution}, the classes are well distributed among the three sets, but they are not at all balanced. The Vehicle and Boat classes represent only less than \SI{3}{\percent} of the surface, while the Building class covers more than \SI{50}{\percent} of the surface.

\subsection{Evaluation method}\label{sec:Evaluation method}

To compare our different sampling methods and the importance of the selected features, we used the rigid KPConv model \citep{thomas2019} to semantize the sampled point clouds, trained during 100 epochs. We chose KPConv because it was among those obtaining better results on the SUM dataset \citep{gao2021} and because the model was already implemented in Torch Points 3D \citep{chaton2020}, the framework we chose to perform the semantic segmentation of our sampled point clouds. We adapted the model to sum the logit (the output of the last layer before the softmax) of each point belonging to the same face and thus obtain a segmentation by face instead of by point, since our mesh is labeled at each face.

As each tile contains on average 4 million points, it is not possible to train on a complete tile at each stage. The model therefore learns at each time from 1024 subsets of points distributed in batches of 16. As recommended by \citet{hu2021}, to draw this subset we choose a point and its $k$ nearest neighbors. This allows to obtain a fixed number of points per batch, whereas this would not have been the case if we had drawn all the points present in a sphere of given size. We tried several values for $k$, which affects the model's receptive field during learning.

As explained in Section \ref{sec:Dataset}, our classes are quite unbalanced. To counteract this, when we draw a subset at random, the center point is not drawn uniformly among all points, but inversely proportional to the frequency of each class. Thus, on average, each class is visited as much as the others. This is only true for the center point of each subset, so our draw is not perfectly balanced, but comes close.

The validation set is used to select the best model obtained over the 10 epochs. The validation metrics are also obtained from a draw of 1024 subsets in each epoch. However, the test operation is performed on all the test set. It is split into subsets of size $k$ which overlap so that all points of the test tiles belong to at least one sphere. The logit obtained at the output of the model are then averaged for each point that belongs to more than one sphere.

\begin{table*}[t]
    \setlength{\tabcolsep}{3.5pt}
	\centering
		\begin{tabular}{l *{13}{c}}
		    \hline
	         & & \multicolumn{6}{c}{IoU} \\
		      & Density & \rotatebox{90}{\parbox{1.5cm}{Terrain}} & \rotatebox{90}{\parbox{1.5cm}{High \\ Vegetation}} & \rotatebox{90}{\parbox{1.5cm}{Building}} & \rotatebox{90}{\parbox{1.5cm}{Water}} & \rotatebox{90}{\parbox{1.5cm}{Vehicle}} & \rotatebox{90}{\parbox{1.5cm}{Boat}} & mIoU & OA & mAcc\\
			& (\si{pts\per\meter\squared}) & (\si{\percent}) & (\si{\percent}) & (\si{\percent}) & (\si{\percent}) & (\si{\percent}) & (\si{\percent}) & (\si{\percent}) & (\si{\percent}) & (\si{\percent}) \\
			\hline
			SUM + KPConv \citep{gao2021} & 10 & 86.5 & 88.4 & 92.7 & 77.7 & 54.3 & 13.3 & 68.8 & 93.3 & 73.7 \\
            SUM \citep{gao2021} & 10 & 83.3 & 90.5 & 92.5 & 86.0 & 37.3 & 7.4 & 66.2 & 93.0 & 70.6 \\
            \hline
            PDS(0.4) with RGB and 10240 pts/sphere & 11.7 & 90.6 & 93.5 & \textbf{96.9} & 96.9 & 71.2 & \textbf{68.0} & \textbf{86.2} & 97.0 & 91.6 \\
            \hline
            With face normal & 11.7 & 91.0 & 94.3 & 96.8 & 96.9 & \textbf{71.8} & 64.7 & 85.9 & \textbf{97.1} & 91.1 \\
            With normal & 11.7 & 89.9 & \textbf{94.4} & 96.4 & 97.0 & 70.5 & 61.6 & 85.0 & 96.9 & 90.9 \\
            With 5120 pts/sphere & 11.7 & 89.0 & 91.8 & 96.1 & 97.2 & 69.6 & 62.4 & 84.3 & 96.6 & \textbf{92.2} \\
            With PDS(0.2) and 20480 pts/sphere & 47.0 & 89.9 & 94.0 & 96.6 & 97.0 & 57.2 & 63.0 & 82.9 & 96.9 & 87.4 \\
            With TS(4.0) & 15.4 & \textbf{91.5} & 93.2 & 96.4 & \textbf{97.9} & 68.7 & 48.5 & 82.7 & 97.0 & 90.7 \\
            With elevation & 11.7 & 85.9 & 91.5 & 94.3 & 96.1 & 55.2 & 47.1 & 78.4 & 95.4 & 81.5 \\
            Without RGB & 11.7 & 86.4 & 88.4 & 94.7 & 91.9 & 59.4 & 34.9 & 76.0 & 95.0 & 84.3 \\
            \hline
		\end{tabular}
	\caption{Comparison of various sampling and feature selection. PDS($r$)~=~Poisson disk sampling, TS($s$)~=~texel sampling.}
\label{tab:results}
\end{table*}

To evaluate the results, we use three classical metrics: Overall Accuracy (OA), Intersection over Union (IoU) and class accuracy (Acc). The first one is defined for each class, while the last two evaluate the whole model. IoU and Acc compensates for different class frequencies as opposed to OA that does not balance different class frequencies giving higher influence on large classes.

If ${n}_{ij}$ is the number of faces from ground-truth class $i$ label as $j$ by our network,
\newline The IoU for class $i$ is:
\begin{equation}\label{equ:1}
	{\textrm{IoU}}_{i}=\frac{n_{ii}}{n_{ii}+\sum_{j\neq i}{n_{ij}}+\sum_{k\neq i}{n_{ki}}}
\end{equation}

The OA is:
\begin{equation}\label{equ:2}
	\textrm{OA}=\frac{\sum_i{n_{ii}}}{\sum_j{\sum_k{n_{jk}}}}
\end{equation}

The Acc for class $i$ is:
\begin{equation}\label{equ:3}
	{\textrm{Acc}}_i=\frac{n_{ii}}{\sum_j{n_{ij}}}
\end{equation}

The average with respect to each class of IoU and Acc is noted mIoU and mAcc.

All our codes are public. The sampling program is available at \href{https://github.com/umrlastig/Mesh-2-Point-Cloud/}{\texttt{github.com/umrlastig/Mesh-2-Point-Cloud}} and the semantization part has been implemented with the Torch Points 3D framework \citep{chaton2020} and is available at \href{https://github.com/umrlastig/torch-points3d/tree/ggrzeczkowicz_Mesh-Segmentation}{\texttt{github.com/umrlastig/torch-points3d}}. Our training statistics and results can be checked on WandB at \href{https://wandb.ai/ggrzeczkowicz/sum-semantic-segmentation/}{\texttt{wandb.ai/ggrzeczkowicz/sum-semantic-segmentation}}.

We performed all our experiments on a computer equipped with 184 GB of RAM, 6 CPUs and a Tesla V100 GPU with 32 GB of RAM. Each training session takes an average of 8 hours.

\subsection{Results}\label{sec:Results}

\begin{figure*}[ht]
\centering
    \subfloat[Ground-truth]{
        \includegraphics[width=0.19\textwidth]{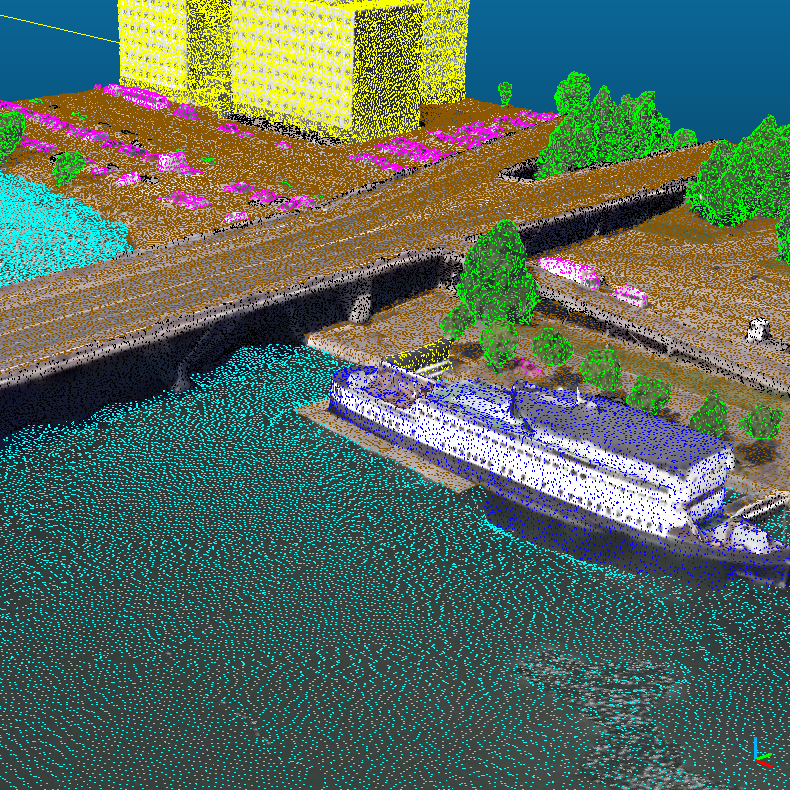}
        \includegraphics[width=0.19\textwidth]{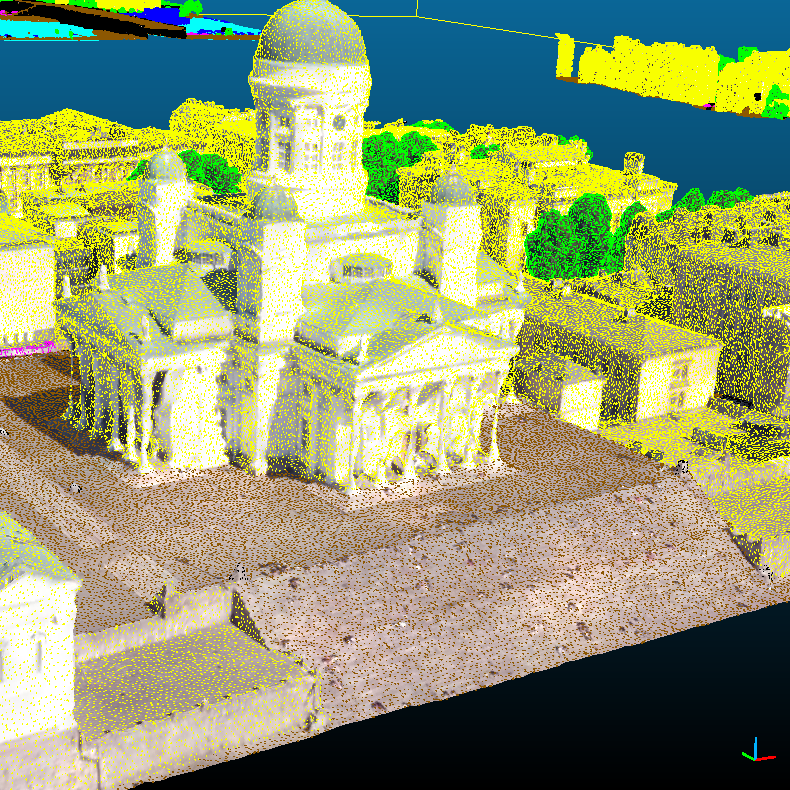}
        \includegraphics[width=0.19\textwidth]{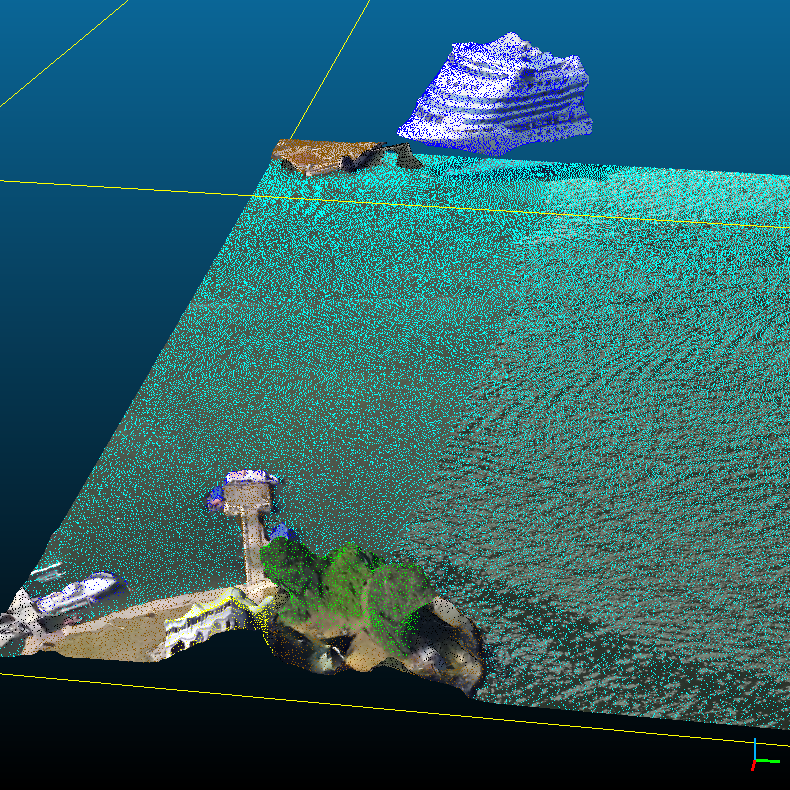}
        \includegraphics[width=0.19\textwidth]{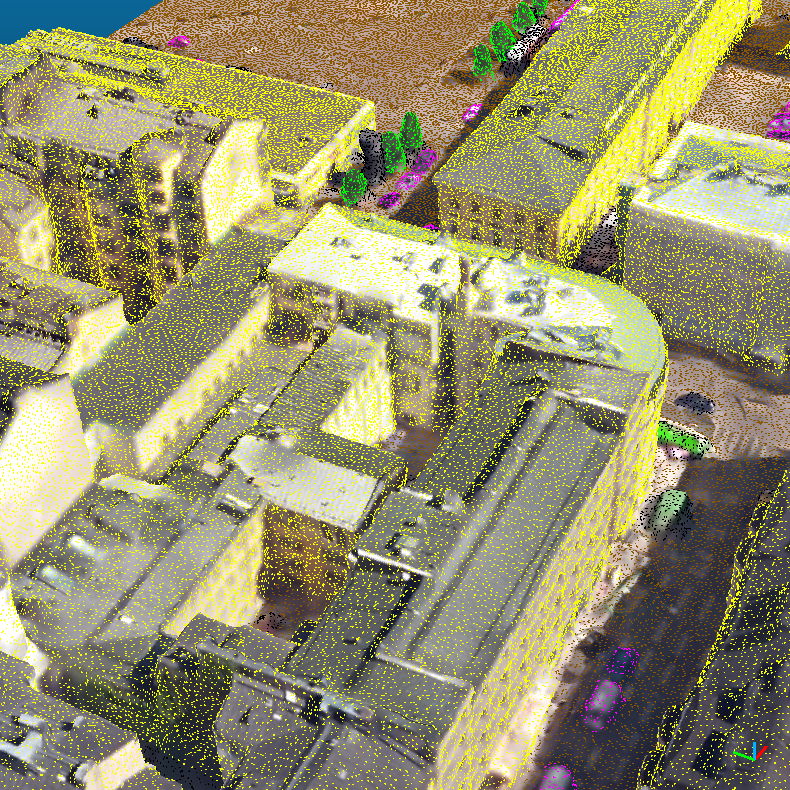}
        \includegraphics[width=0.19\textwidth]{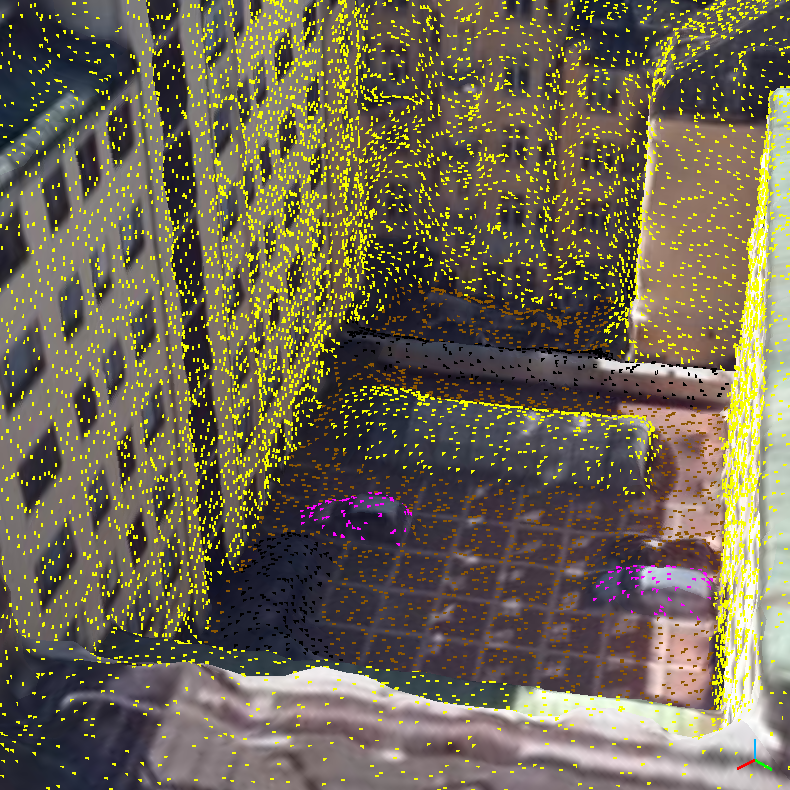}
        \label{subfig:result-gt}
    }
    \hspace{0pt}
    \subfloat[Predictions]{
        \includegraphics[width=0.19\textwidth]{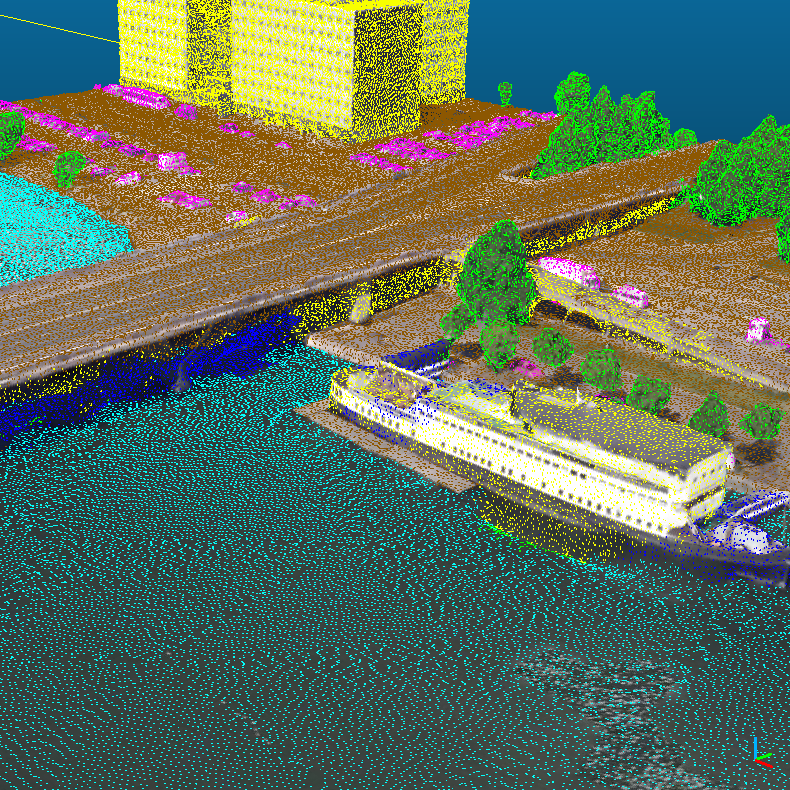}
        \includegraphics[width=0.19\textwidth]{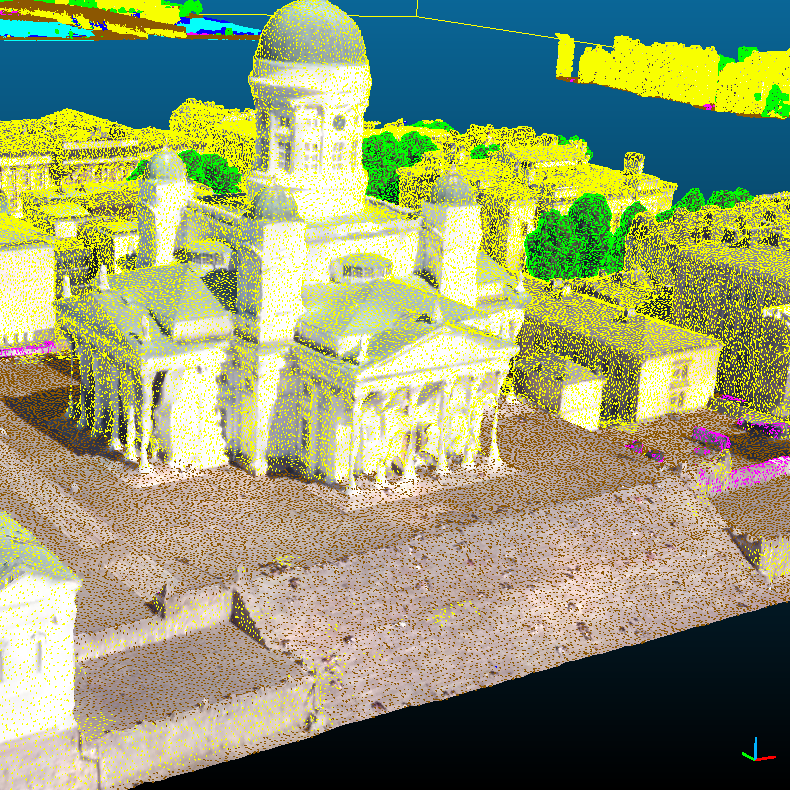}
        \includegraphics[width=0.19\textwidth]{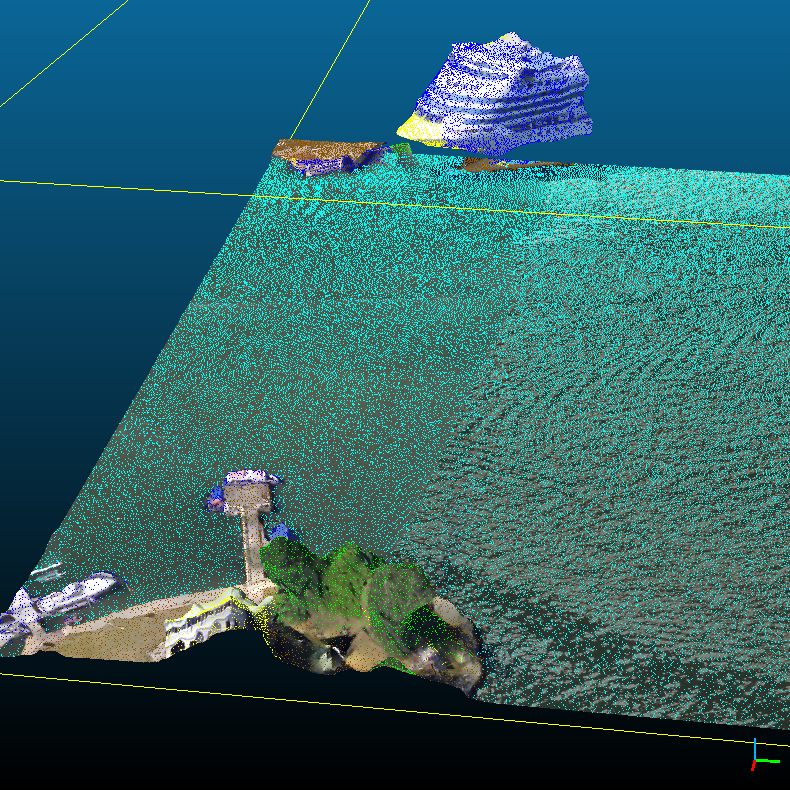}
        \includegraphics[width=0.19\textwidth]{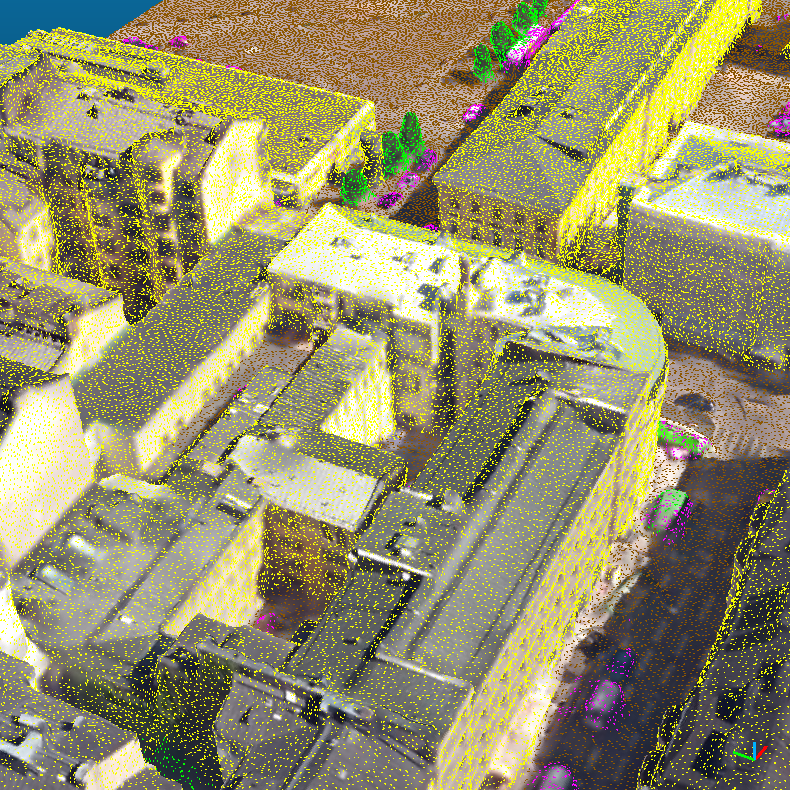}
        \includegraphics[width=0.19\textwidth]{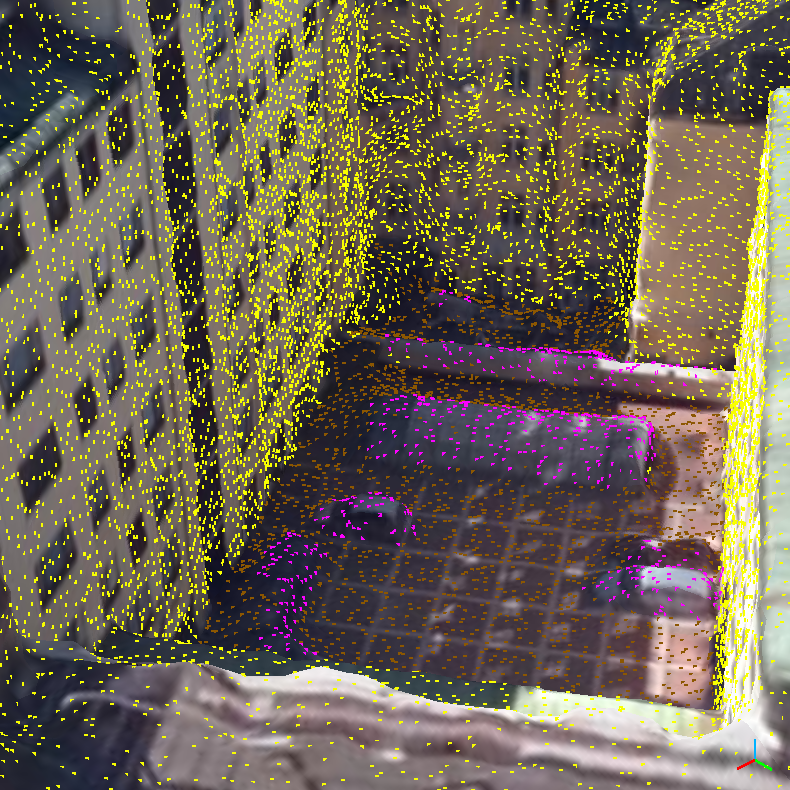}
        \label{subfig:result-prediction}
    }
    \hspace{0pt}
    \subfloat[Error map]{
        \includegraphics[width=0.19\textwidth]{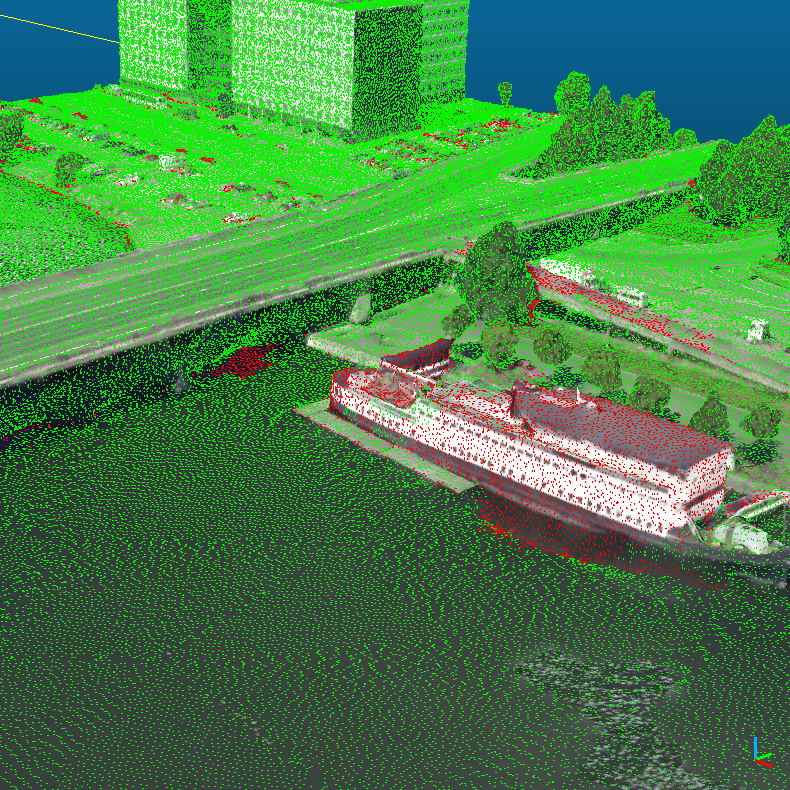}
        \includegraphics[width=0.19\textwidth]{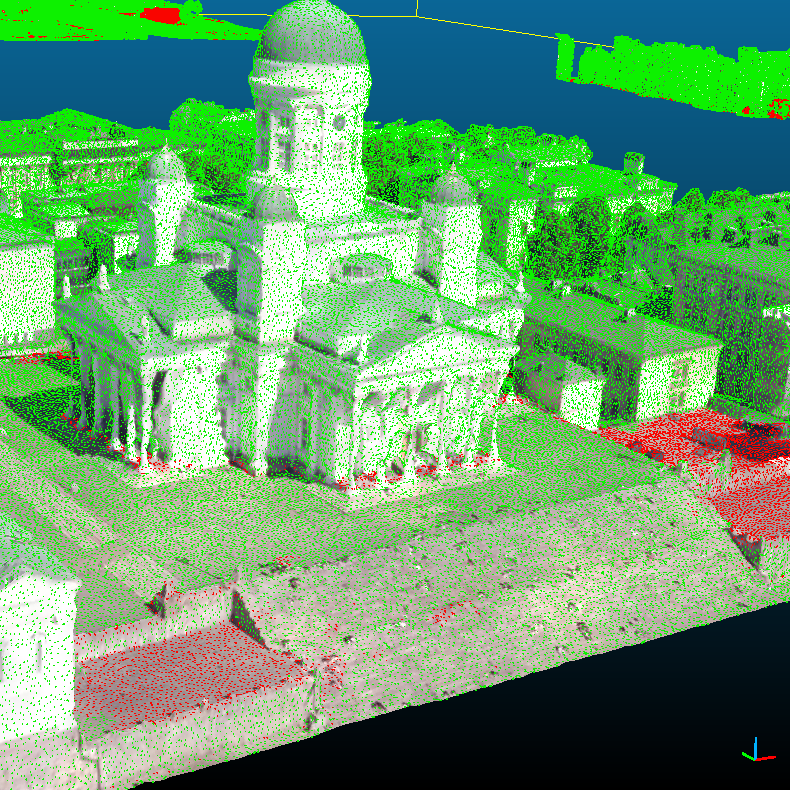}
        \includegraphics[width=0.19\textwidth]{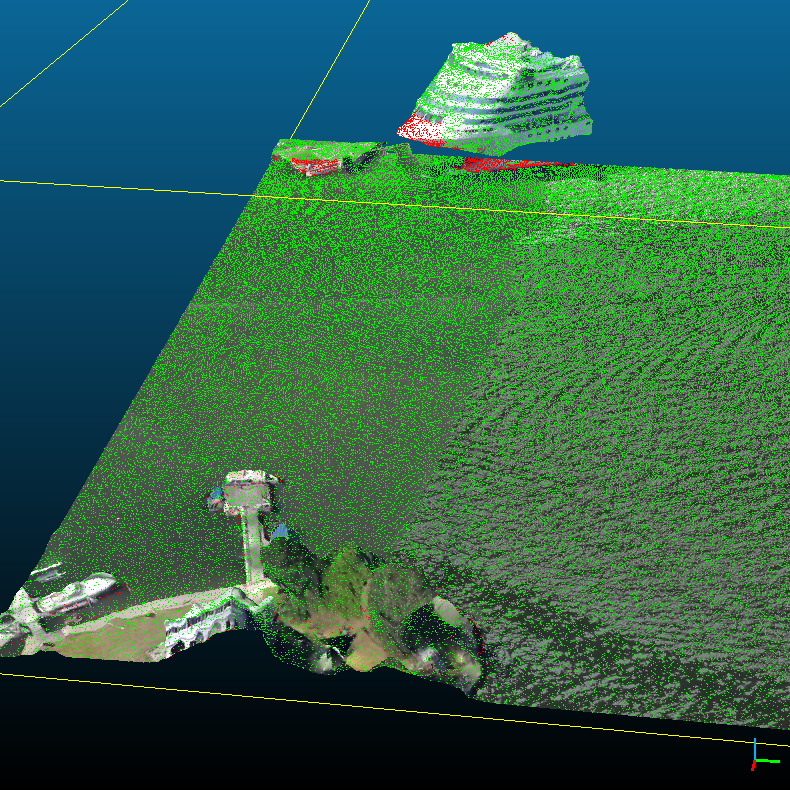}
        \includegraphics[width=0.19\textwidth]{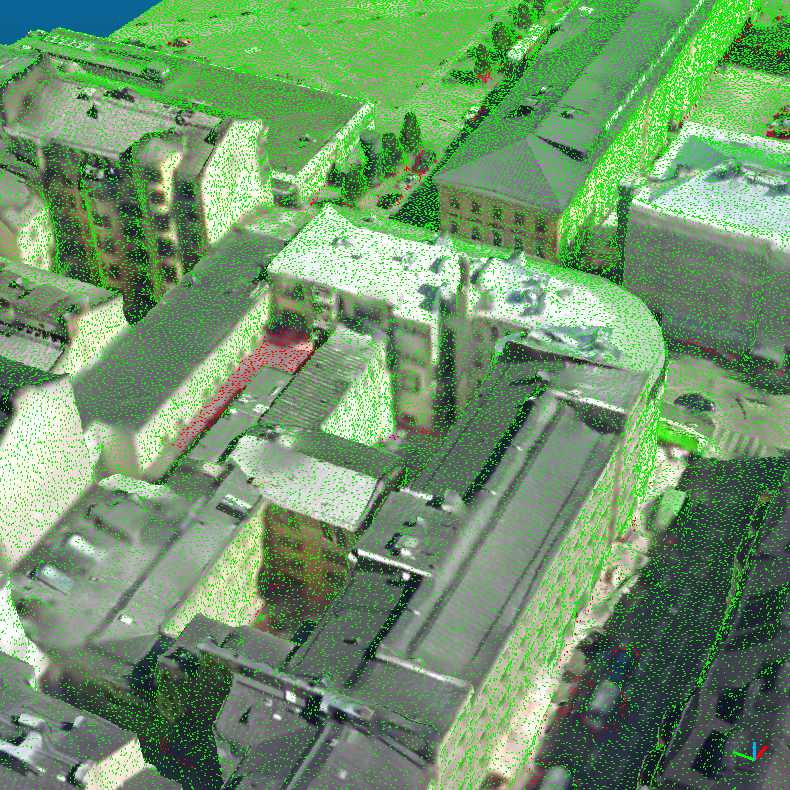}
        \includegraphics[width=0.19\textwidth]{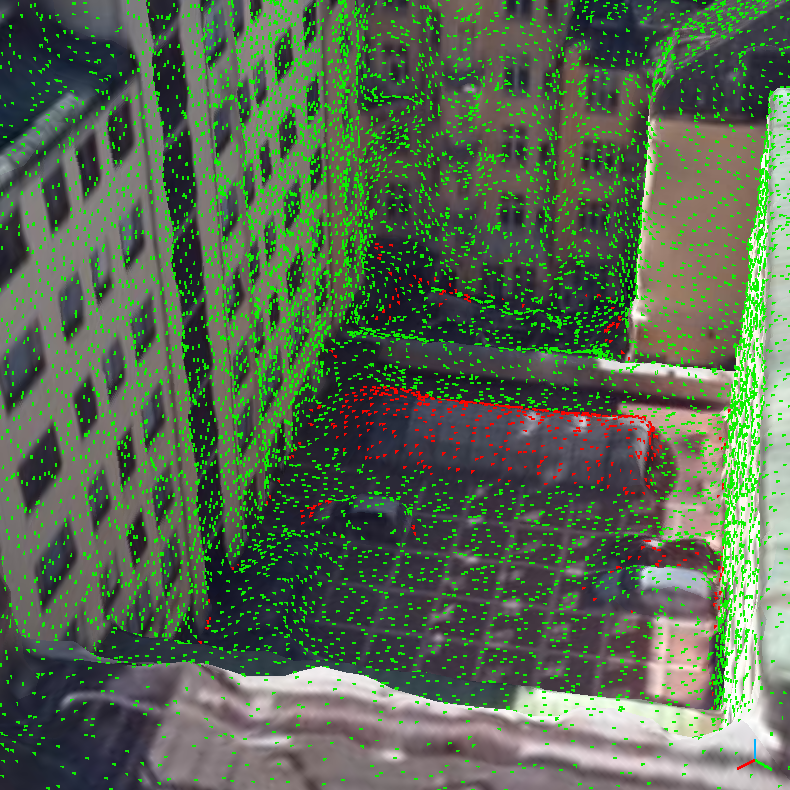}
        \label{subfig:result-error-map}
    }
\caption{Results obtain with Poisson disk sampling, $r = 0.4$, RGB and 10240 pts/sphere.}
\label{fig:results}
\end{figure*}

We obtain a maximum mIoU with a Poisson disk sampling, $r = 0.4$, only RGB as feature and 10240 points per subset. In our results, Vehicle and boats are the main sources of errors as they are much less represented in the dataset and are smaller objects that suffer more from under-sampling. While water is also poorly represented, it is a much more identifiable object, which explains the much better results. From this result we conduct an ablation study (Table \ref{tab:results}) to quantify the influence of each parameter.

We observe that the most important feature is the color whose absence greatly lowers the quality of the results. This result is quite logical as some surfaces such as water and ground are distinguishable only by radiometric features. The contribution of the elevation seems to decrease the performance, especially on small objects such as vehicles and boats. This can be caused by several phenomena. First, as we discuss in \ref{sec:Feature selection}, the elevation we have chosen is the simplest, but not necessarily the most rigorous. Also, when we look at the training statistics, the elevation seems to confuse our network more than anything else, causing our loss to decrease much more slowly. In the future, it might be worth trying to integrate a true DTM, as related work on urban classification \citep{rouhani2017,tutzauer2019,laupheimer2020} has shown that elevation is generally among the most important features for classification. The contribution of the normals is marginal, they improve the IoU of some classes while decreasing that of others, causing a slight decrease in mIoU. 

Reducing the number of points per sphere by two does not seem to decrease the quality of the results significantly (about two IoU points in each class) while it allows to divide the learning time by two (4 hours instead of 8). This could be interesting for GPUs with less RAM. However, it should be noted that this increases the number of sub-samples to be classified in the test phase. A priori, increasing the density of the point cloud does not lead to better results either. However, we must take into account that it is the receptive field of each point that could be decisive, and that despite the fact that we have doubled the size of the subsamples, it is by $2^3=8$ that we must multiply the size of the subsamples to keep a receptive field of the same dimensions when we divide by two the distance between the sampling points. For a comparable density, texel sampling seems to obtain results of the same order of magnitude, but on average slightly worse. This sampling being slightly less uniform than Poisson disk sampling and anisotropic, it could introduce a slight bias that degrades the results. Grid subsampling does not have a measurable effect, at equal density, in our results.

We also compare our results to \citet{gao2021} who established a first benchmark of their dataset. They used the same algorithm as for their automatic labeling and also performed a point sampling and then used different semantic segmentation algorithms. In this benchmark, KPConv obtains the best results (which partly explains our choice to use this method), but their results are far below the ones that we obtain (cf Table \ref{tab:results}). As this benchmark uses a comparable density of points (\SI{10}{pts\per\metre\squared}) and also Poisson disk sampling, the difference may be explained by a different setting of the semantization algorithm as the chosen hyper-parameters are not publicly available. Their results on the boat class may also suggest an unresolved class imbalance problem.

The qualitative results (Figure \ref{fig:results}) give us a better idea of the sources of error. The errors appear at the interface between two classes. It should be noted that the ground truth is labeled at the face level, but sometimes the boundary between two classes is not exactly at the interface between two faces, which can explain very slight differences. We see here an advantage of classifying at the point level, allowing to classify the mesh at the texel level and not at the face level.  Sometimes it is a real error and sometimes the labeling is more delicate. In the first example, a boat that is too close to the water is classified as a building by mistake. In the second example, the terrace of the monument is labeled as a building when it is classified as a ground. In the third example, there is less error than in the first example with the boat, but there are still some concerns. The fourth example shows us a roof falsely classified as ground, which makes explicit the importance of having a good elevation feature. Finally, the last example is a tent in a parking lot that has been classified as a vehicle.

\section{CONCLUSION AND PERSPECTIVES}\label{sec:PERSPECTIVES}

\sloppy

Textured meshes are becoming an increasingly popular representation combining the 3D geometry and radiometry of real scenes. When analysing the underlying scene, and in particular its semantics, it is still an open question to know if the best strategy is to semantize the pixels of the input images, the 3D points of the photogrammetric point cloud, both at the same time in a multi-modal approach \citep{robert2022}, or through the textured mesh as proposed in this paper. However we believe that the raw (images) or intermediate (point clouds) products might often be lost or simply discarded by the production pipeline, such that being able to semantize the textured mesh directly has an inherent benefit.

We investigated other alternatives to textured mesh semantic segmentation, in particular leveraging graph convolutions on the primal or dual structure of the mesh, but the patch structure of the texture makes it always complicated to integrate textural information, more accurate than just the average texel of the face, at node level and our best effort could not reach the same level of quality as with the much simpler sampling approach. The main perspective of this work is to extend it to instance segmentation (separating instances of adjacent objects such as buildings and vehicles) and structured reconstruction (going from a textured mesh to a LoD2 city model).

{
	\begin{spacing}{1} 
		\normalsize
		\bibliography{bibliography} 
	\end{spacing}
}

\end{document}